\documentclass[nohyperref]{article}

\usepackage{amsmath,amsfonts,bm}

\def\eqref#1{equation~\ref{#1}}

\def\1{\bm{1}}

\def\vx{{\bm{x}}}

\DeclareMathAlphabet{\mathsfit}{\encodingdefault}{\sfdefault}{m}{sl}
\SetMathAlphabet{\mathsfit}{bold}{\encodingdefault}{\sfdefault}{bx}{n}

\usepackage{microtype}
\usepackage{graphicx}
\usepackage{subfigure}
\usepackage{booktabs} %

\usepackage{hyperref}

\usepackage[accepted]{icml2023}

\usepackage{amsmath}
\usepackage{amssymb}
\usepackage{mathtools}
\usepackage{amsthm}
\usepackage{pifont}
\newcommand{\xmark}{\ding{55}}%

\usepackage[capitalize,noabbrev]{cleveref}

\theoremstyle{plain}

\theoremstyle{definition}

\theoremstyle{remark}

\usepackage[textsize=tiny]{todonotes}

\usepackage{booktabs,makecell,multirow,hhline,threeparttable} 
\usepackage{xcolor,colortbl,pgf}
\definecolor{Color1}{RGB}{240, 240, 240}
\usepackage{makecell}
\usepackage{xspace}

\usepackage[algo2e]{algorithm2e} 

\SetCommentSty{mycommfont}
\usepackage{pifont}%
\newcommand{\cmark}{\ding{51}}%

\newcommand{\HSIC}{\operatorname{HSIC}}
\newcommand{\method}{DualHSIC\xspace}
\newcommand{\methodabbr}{DualHSIC\xspace}
\newcommand{\ie}{\textit{i.e.\xspace}}

\newcommand{\bottleneck}{HBR\xspace}
\newcommand{\bottleneckfull}{HSIC-Bottleneck for Rehearsal\xspace}
\newcommand{\interaction}{HA\xspace}
\newcommand{\interactionfull}{HSIC Alignment\xspace}
\newcommand{\zheng}[1]{\textcolor{black}{#1}}
\newcommand{\zifeng}[1]{\textcolor{black}{#1}}
\icmltitlerunning{DualHSIC: HSIC-Bottleneck and Alignment for Continual Learning}

\begin{document}

\twocolumn[
\icmltitle{DualHSIC: HSIC-Bottleneck and Alignment for Continual Learning}

\icmlsetsymbol{equal}{*}

\begin{icmlauthorlist}
\icmlauthor{Zifeng Wang}{equal,yyy}
\icmlauthor{Zheng Zhan}{equal,yyy}
\icmlauthor{Yifan Gong}{yyy}
\icmlauthor{Yucai Shao}{comp}
\icmlauthor{Stratis Ioannidis}{yyy}
\icmlauthor{Yanzhi Wang}{yyy}
\icmlauthor{Jennifer Dy}{yyy}
\end{icmlauthorlist}

\icmlaffiliation{yyy}{Northeastern University}
\icmlaffiliation{comp}{University of California, Los Angeles}

\icmlcorrespondingauthor{Zifeng Wang}{zifengwang@ece.neu.edu}
\icmlcorrespondingauthor{Zheng Zhan}{zhan.zhe@northeastern.edu}

\icmlkeywords{Machine Learning, ICML}

\vskip 0.3in
]

\printAffiliationsAndNotice{\icmlEqualContribution} %

\begin{abstract}
Rehearsal-based approaches are a mainstay of continual learning (CL). They mitigate the catastrophic forgetting problem by maintaining a small fixed-size buffer with a subset of data from past tasks. While most rehearsal-based approaches study how to effectively exploit the knowledge from the buffered past data, little attention is paid to the inter-task relationships with the critical task-specific and task-invariant knowledge. By appropriately leveraging inter-task relationships, we propose a novel CL method named \textit{\methodabbr} to boost the performance of existing rehearsal-based methods in a simple yet effective way. \methodabbr consists of two complementary components that stem from the so-called Hilbert Schmidt independence criterion (HSIC): \textit{\bottleneckfull} (\bottleneck) lessens the inter-task interference and \textit{\interactionfull} (\interaction) promotes task-invariant knowledge sharing. Extensive experiments show that \methodabbr can be seamlessly plugged into existing rehearsal-based methods for consistent performance improvements, and also outperforms recent state-of-the-art regularization-enhanced rehearsal methods. Source code will be released. 
\end{abstract}

\section{Introduction}

Continual learning (CL) aims at enabling a single model to learn a sequence of tasks without \textit{catastrophic forgetting}~\citep{mccloskey1989catastrophic} - the central problem of CL that models are prone to performance deterioration on previously seen tasks. A large body of work attempts to address CL from different perspectives~\citep{kirkpatrick2017overcoming,mallya2018packnet,aljundi2018memory}. \textit{Rehearsal-based} methods~\cite{aljundi2018memory, chaudhry2019tiny, buzzega2020dark} have gained popularity due to their simplicity, effectiveness, and generality.

\begin{figure}[t]
    \centering
    \includegraphics[width=0.89\columnwidth]{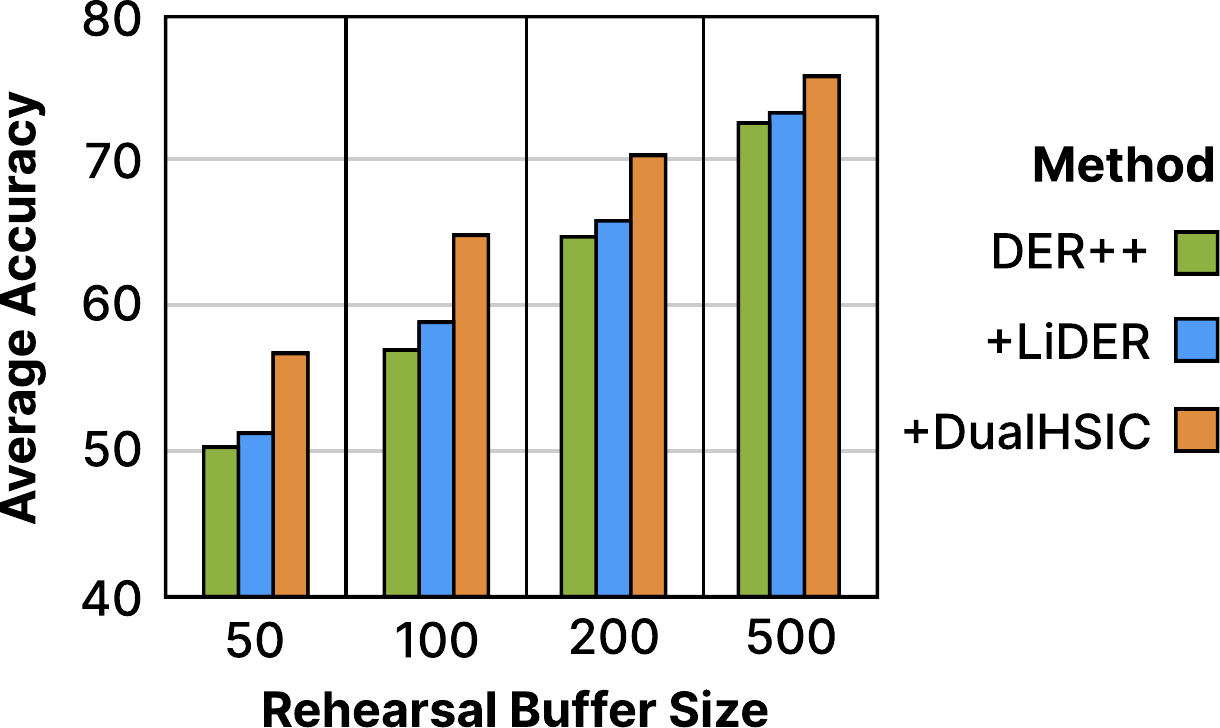}
    \caption{Comparison between \methodabbr, and two representative SOTA CL methods, LiDER and DER++. Both \methodabbr and LiDER are built upon DER++. \methodabbr achieves especially better performance with smaller buffer sizes.}
    \label{fig:intro}
\end{figure}
The core idea of rehearsal is to maintain a small fix-sized memory buffer to save a subset of data from past tasks. When training on the current task, the model also revisits the buffered data to consolidate learned knowledge. Given the limited buffer size, it is challenging to keep generally discriminative representations for old tasks, because of overfitting~\cite{verwimp2021rehearsal}. Existing methods mainly focus on data augmentation~\cite{buzzega2020dark, cha2021co2l} and importance-based buffer example selection~\cite{aljundi2019gradient,yoon2021online}.

Despite state-of-the-art (SOTA) performance, these approaches mostly consider how to better exploit knowledge from the buffered past data. It is notable that the inter-task relationship is also important yet under-investigated in rehearsal-based work: \textit{How does learning the current task affect the consolidation of past knowledge?} To answer this question, we are inspired by the Complementary Learning Systems (CLS)~\cite{kumaran2016learning, mcclelland1995there} theory and CLS-based CL methods~\cite{pham2021dualnet, wang2022dualprompt}, which suggest that \textit{task-specific} and \textit{task-invariant} knowledge are critical for CL. Therefore, when learning the current task, we have to prevent the current task-specific knowledge from interfering with past knowledge, and leverage task-invariant knowledge to better consolidate the past.

A straightforward approach is to directly combine CLS-based methods with a rehearsal buffer. Although such an approach has proved effective, major components such as an additional backbone~\citep{pham2021dualnet}, prompting mechanisms~\citep{wang2022dualprompt}, or re-design of the rehearsal buffer system~\cite{arani2022learning} are required. Ideally, one would prefer a more general method that can be seamlessly incorporated into most existing rehearsal-based methods with minimal tweaks.

To this end, we propose a novel CL method, \textit{\methodabbr}, that  improves existing rehearsal-based methods from a unique perspective: leveraging the so-called Hilbert-Schmidt independence criterion (HSIC)~\cite{gretton2005measuring} to learn better feature representations for rehearsal. HSIC is much more tractable and efficient than mutual information to measure statistical independence, and has been widely adopted for various sub-fields in machine learning~\cite{wang2020open, ma2020hsic}. \methodabbr consists of two complementary components: 1) \textit{\bottleneckfull} (\bottleneck) that mitigates inter-task interference by removing uninformative task-specific knowledge introduced by learning on the current task from the buffered data, such that inter-task interference and catastrophic forgetting are mitigated; 2) \textit{\interactionfull} (\interaction) that encourages task-invariant knowledge sharing between current and past tasks for positive knowledge transfer~\citep{hadsell2020embracing}. These components can be easily plugged into existing rehearsal-based methods with consistent performance improvement. \zifeng{In Figure~\ref{fig:intro}, we demonstrate that \methodabbr outperforms advanced SOTA methods under different buffer sizes, with especially larger margins at small buffer sizes.} 

To further demonstrate the generality and effectiveness of \methodabbr, we also conduct comprehensive experiments on multiple CL benchmarks. We show that \methodabbr works collaboratively with various existing rehearsal-based CL methods, leading to consistent improvement upon SOTA results. We also conduct in-depth exploratory experiments to analyze the effectiveness of core designs of \methodabbr.

Overall, our work makes the following contributions:
\begin{itemize}
    \item We propose \methodabbr, a general CL method that improves a wide spectrum of rehearsal-based methods. \methodabbr mitigates inter-task interference and encourages task-invariant knowledge sharing between tasks via the novel \bottleneck and \interaction losses.
    \item Comprehensive experiments demonstrate that \methodabbr consistently improves SOTA rehearsal-based methods by at most $7.6\%$, and also outperforms stronger regularization-enhanced methods by at most $6.5\%$.
    \item To the best of our knowledge, our work is the first to bring HSIC to CL to learn better representations in a systematic way.
\end{itemize}

\section{Related Work}
\textbf{Continual Learning.} Existing CL works can be mainly categorized into regularization-based, architecture-based, and rehearsal-based approaches. Regularization-based approaches \cite{kirkpatrick2017overcoming, zenke2017continual, li2017learning, aljundi2018memory} introduce additional terms in the loss function to penalize the model change on important weights for the purpose of protecting earlier tasks. Architecture-based approaches \cite{rusu2016progressive, mallya2018packnet, wang2020learn, wang2021learning, yan2021dynamically} dynamically expand the model capacity or isolate existing model weights to reduce the interference between the new tasks and the old ones. Rehearsal-based approaches allow access to a memory buffer with examples from prior tasks and train the model jointly with the current task. With its simplicity and efficacy, the idea of rehearsal enjoys great popularity and  has been adopted by many state-of-the-art methods~\cite{buzzega2020dark, cha2021co2l, pham2021dualnet, wang2022sparcl}. In this work, we present \methodabbr as a general surrogate loss that improves rehearsal-based methods.

\textbf{Hilbert Schmidt Independence Criterion (HSIC).} As a statistical dependency measure, HSIC~\cite{gretton2005measuring} has been widely applied in various machine learning applications, such as dimensionality reduction~\cite{niu2011dimensionality}, 
clustering~\cite{wu2020deep}, feature selection~\cite{song2012feature}, and class discovery~\cite{wang2020open}. HSIC captures non-linear dependencies between random variables and has the advantage of easy empirical estimation over mutual information (MI). Recently,~\citet{ma2020hsic} propose the HSIC-bottleneck as an alternative for cross-entropy loss.~\citet{wang2021revisiting} and~\citet{jian2022pruning} further demonstrate how HSIC-bottleneck strengthens a model's adversarial robustness. However, no prior work has studied the application of HSIC under the context of CL. As a very first attempt, we propose two novel complementary HSIC-related losses that address catastrophic forgetting from a unique perspective.

Among the latest CL works, OCM~\citep{guo2022online} and LiDER~\citep{bonicelli2022effectiveness} are the closest to our work, in terms of the common target to improve rehearsal-based methods via surrogate loss terms. However, we would still like to emphasize that our work is different and novel. OCM proposes an MI-based loss through a complicated contrastive learning proxy~\cite{oord2018representation}, while our work introduces two different HSIC losses with a simple empirical evaluation strategy. LiDER constrains the Lipschitz constant of a model to strengthen the robustness of the decision boundary, while~\methodabbr has a totally different motivation and methodology. Moreover, different from both works, we explicitly consider the inter-task relationship in CL. We also show that~\methodabbr outperforms OCM and LiDER consistently in practice (Table~\ref{tab:table2}).

\section{Preliminaries}

\subsection{Continual Learning Problem Setting}
In supervised continual learning, a sequence of tasks $\mathcal{D} = \{\mathcal{D}_1, \ldots, \mathcal{D}_T\}$ arrive in a streaming fashion, where each task $\mathcal{D}_t = \{(\vx^t_{i}, y^t_{i})\}_{i=1}^{n_t}$ contains a separate target dataset, \ie, $\mathcal{D}_i \cap \mathcal{D}_j = \emptyset$. A single model needs to adapt to them sequentially, with only access to $\mathcal{D}_t$ at the $t$-th task. In practice, we allow a small fix-sized rehearsal buffer $\mathcal{M}$ to save data from past tasks. At test time, we mainly focus on one of the most challenging class-incremental (Class-IL) setting, where no task identity is available for the coming test examples.

In general, given a prediction model $h_\theta$ parameterized by $\theta$, a large body of continual learning work seeks to optimize for the following loss at the $t$-th task:
\begin{align} \small
\label{eq:original_loss}
\begin{split}
\mathcal{L}_{\text{CL}}(\theta) = \sum_{\vx, y \in \mathcal{D}_t} \ell(h_{\theta}(\vx), y) + \sum_{\vx^M, y^M \in \mathcal{M}} \ell_{\mathcal{M}}(h_{\theta}(\vx^M), y^M),
\end{split}
\end{align}
where $\ell$ and $\ell_{\mathcal{M}}$ are losses for the current data and buffered data, respectively. For example, \citet{chaudhry2019tiny} applies cross-entropy as both losses, while many recent works present different losses or add additional auxiliary loss terms~\cite{buzzega2020dark, cha2021co2l}. Although $\mathcal{L}_{\text{CL}}$ can take many forms depending on the actual method, our method~\methodabbr presents a model-agnostic loss that can be plugged into most rehearsal-based CL methods to improve the overall performance.

\subsection{Hilbert-Schmidt Independence Criterion}
The Hilbert-Schmidt independence criterion (HSIC) is a statistical measure for identifying dependencies between two random variables, which was first introduced by \citet{gretton2005measuring}. HSIC calculates the Hilbert-Schmidt norm of the cross-covariance operator of the distributions in the Reproducing Kernel Hilbert Space (RKHS). Similar to the widely used Mutual Information~\cite{shannon1948mathematical}, HSIC is able to detect non-linear dependencies with the advantage of easy empirical estimation over MI.

Given two random variables $X$ and $Y$, the HSIC between them is formally defined as:
\begin{align}\label{eq:hsic}
\begin{split}
\HSIC&(X, Y) 
= \mathbb{E}_{X Y X^{\prime} Y^{\prime}}\left[k_{X}\left(X, X^{\prime}\right) k_{Y^{\prime}}\left(Y, Y^{\prime}\right)\right] \\
&+\mathbb{E}_{X X^{\prime}}\left[k_{X}\left(X, X^{\prime}\right)\right] \mathbb{E}_{YY^{\prime}}\left[k_{Y}\left(Y, Y^{\prime}\right)\right] \\
&-2 \mathbb{E}_{X Y}\left[\mathbb{E}_{X^{\prime}}\left[k_{X}\left(X, X^{\prime}\right)\right] \mathbb{E}_{Y^{\prime}}\left[k_{Y}\left(Y, Y^{\prime}\right)\right]\right],
\end{split}
\end{align}
where $X'$, $Y'$ are independent copies of $X$, $Y$, respectively, and $k_{X}$, $k_{Y}$ are corresponding kernel functions.

HSIC can be easily approximated empirically without knowing the analytical form of distribution $P_{XY}$. Given $n$ i.i.d. examples $\{(\vx_i, y_i)\}_{i=1}^{n}$ sampled from $P_{XY}$, the empirical estimation of HSIC is:
\begin{equation}
    \label{eq:empirical_hsic}
    {\HSIC}_{e}(X, Y)={(n-1)^{-2}} \operatorname{tr}\left(K_{X} H K_{Y} H\right),
\end{equation}
where $K_X$ and $K_Y$ are kernel matrices with $K_{X_{ij}}=k_{X}(\vx_i, \vx_j)$ and $K_{Y_{ij}}=k_{Y}(y_i, y_j)$, respectively, $\operatorname{tr}(\cdot)$ is the trace operator, and $H = \mathbf{I}- \frac{1}{n} \mathbf{1} \mathbf{1}^\top$ is a centering matrix. In our experiments, we evaluate HSIC terms via this empirical estimation.

\begin{figure}[t]
    \centering
    \includegraphics[width=0.99\linewidth]{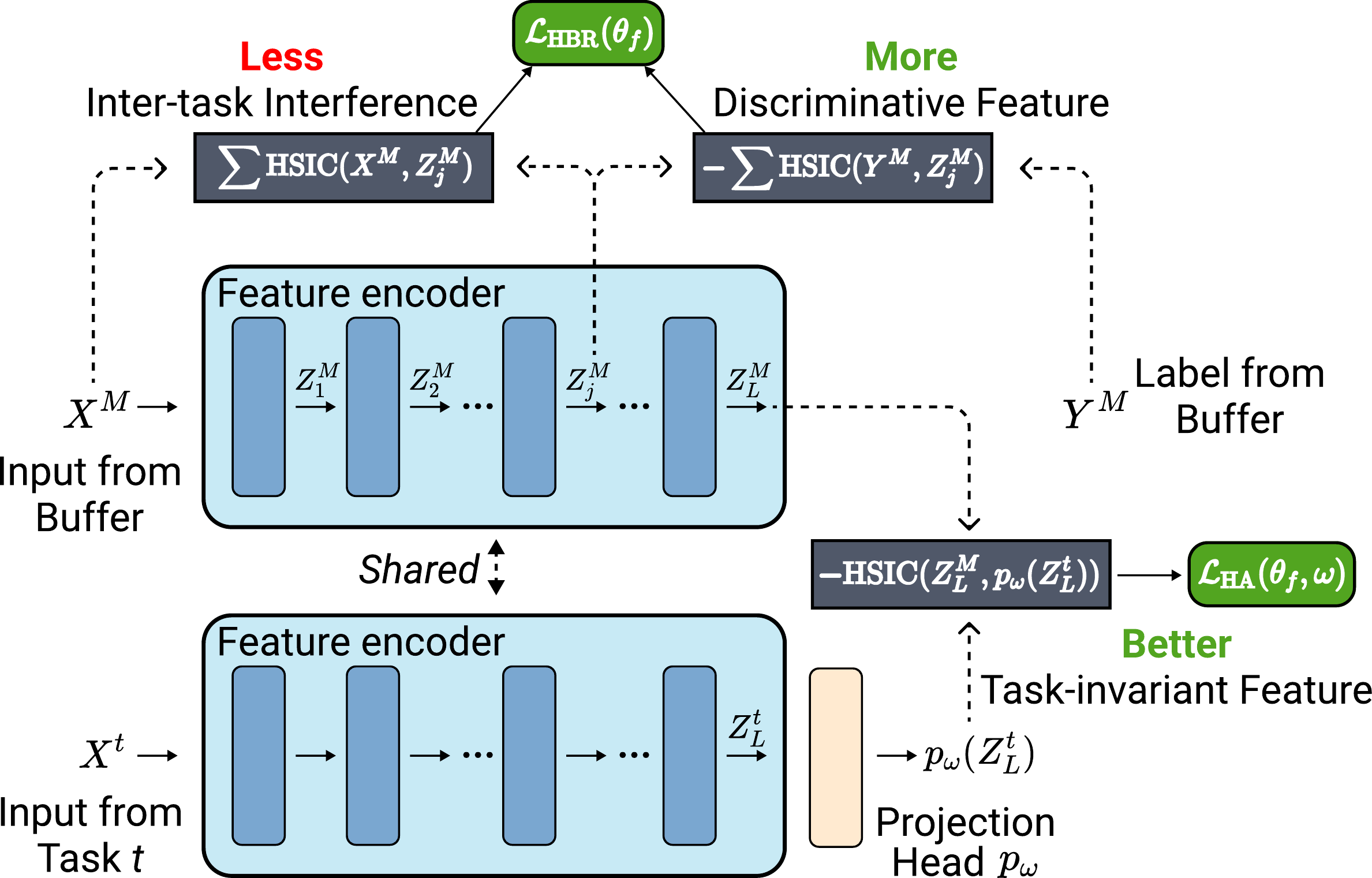}
    \caption{Overview of \methodabbr. \bottleneck is calculated on the buffered data in a \emph{multi-layer} fashion. However, only a sub-component of \bottleneck at the $j$-th intermediate layer is illustrated in the figure for visual clarity. \interaction is calculated between the latent representations of buffered and current data. Both inputs share the same feature encoder and an MLP projection head is used to create an alternative view. Note that the actual \interaction is symmetric, exchanging two input branches in the figure provides the second half of the loss.}
    \label{fig:overview}
\end{figure}

\section{\methodabbr}
In this section, we will present \method, a general continual learning objective that is orthogonal to the existing rehearsal-based framework. As shown in Figure~\ref{fig:overview}, \methodabbr consists of two complementary losses: \emph{\bottleneckfull}, which mitigates inter-task interference, and \emph{\interactionfull} which encourages the sharing of task-invariant knowledge between tasks. %

\subsection{HSIC Bottleneck for Rehearsal} \label{sec:HBR}
During the continual learning process, compared with abundant data from the current task, we only have very limited buffered data from past tasks. This data-imbalance~\cite{hadsell2020embracing, mai2021online} issue makes the model over-focus on task-specific knowledge of the current task, leading to performance deterioration on the past tasks. To address the problem, we propose \bottleneckfull (\bottleneck), a general loss to mitigate inter-task interference and retrain good feature representations for data saved in the rehearsal buffer.

We denote the model, a multi-layer feedforward neural network with $L$ intermediate layers, used in the CL process by $h_\theta: \mathbb{R}^{d_x} \to \mathbb{R}^C$, where $d_x$ is the input dimension and $C$ is the total number of classes of interest during CL. We also decompose $h_{\theta} = g_{\theta_g} \circ f_{\theta_f}$ into the final classification layer $g_{\theta_g}$ and the feature encoder $f_{\theta_f}$ for notation convenience. We use $X^M$ to denote the random variable that represents data saved in the rehearsal buffer $\mathcal{M}$, we further denote by $Z^M_j \in \mathbb{R}^{d_{z_j}}$ its corresponding output of the $j$-th intermediate layer. 
\textit{\bottleneckfull} (\bottleneck) is defined as a penalty loss on the buffered data:
\begin{equation} \label{eq:HBR}
\resizebox{.99\hsize}{!}{$
 \mathcal{L}_{\text{\bottleneck}}(\theta_f) = \lambda_{x} \sum_{j=1}^{L} \HSIC(X^M, Z^M_j) -\lambda_{y} \sum_{j=1}^{L} \HSIC(Y^M, Z^M_j),$}
\end{equation}
where $\lambda_x$ and $\lambda_y$ are balancing coefficients.

\textbf{Mitigating Inter-Task Interference.} Intuitively, minimizing the HSIC between $X^M$ and $Z^M_j$ aims at reducing the noisy information contained within the latent representation $Z^M_j$ w.r.t. the input $X^M$. When learning the current task $t$, the model undoubtedly extracts useful knowledge from the current data. However, such task-specific knowledge may be uninformative or noisy for past tasks~\cite{ebrahimi2020adversarial, pham2021dualnet, wang2022dualprompt}. Therefore, by only applying the bottleneck loss to the buffered data, we implicitly mitigate the interference from learning task-specific knowledge for the current task to past tasks. \\

\textbf{Maintaining Discriminative Knowledge.} HBR also tries to maximize the HSIC between $Y^M$ and $Z^M_j$, which naturally retains the discriminative information useful for classification. Although it serves a similar purpose as the cross-entropy loss for classification, \citet{wang2021revisiting} demonstrate the necessity of this term in the bottleneck loss empirically; we also verify this observation in our ablation study (Section~\ref{sec:ablation}).\\

\textbf{Asynchronous Consolidation.} Note that we only apply \bottleneck to the buffered data, instead of both buffered and current data. \zifeng{Empirically, we observe that adding the term to both data does not lead to performance improvement (Appendix~\ref{app:cur_data})}; similar results have also been observed by~\citet{bonicelli2022effectiveness}. Intuitively, we already have abundant data for the current task compared to the buffered data, so the learning of the current task is much less prone to inter-task interference. On the other hand, our proposed scheme naturally consolidates knowledge in an asynchronous way to address the stability-plasticity dilemma for continual learners~\citep{abraham2005memory, mermillod2013stability}: the model first learns the current task without \bottleneck for maximum \emph{plasticity}, then rehearse the learned task in future tasks with \bottleneck to maintain \emph{stability}.

\textbf{Alternative Perspective via Robustness}. Although not used in the context of CL, HSIC-bottleneck has been proved by~\citet{wang2021revisiting} both empirically and theoretically to improve the adversarial robustness of the model. Interestingly, ~\citet{bonicelli2022effectiveness} also demonstrate that a more adversarially robust model on the buffered data prevents the decision boundary from eroding, thus mitigating catastrophic forgetting. In this respect, \bottleneck provides another bridge that links adversarial robustness with catastrophic forgetting. 

\begin{algorithm}[t]
\SetAlgoLined
\SetNoFillComment
\textbf{Input}: Model $h_{\theta}$ with $L$-layer feature encoder $f_{\theta_f}$ and classifier $g_{\theta_g}$, projection head $p_\omega$, number of tasks $T$, training epochs of the $t$-th task $K_t$, mini-batch size $B$.\\
\textbf{Initialize:} $\theta$ ($\theta_f$ and $\theta_g$), $p_\omega$ \\
\For{$t = 1,\ldots,T$}{
\For{$e = 1,\ldots,K_t$}{
    Draw a mini-batch $\{(\vx^t_{i}, y^t_{i})\}_{i=1}^{B}$ from current task\\
    Draw a mini-batch $\{(\vx^M_{i}, y^M_{i})\}_{i=1}^{B}$ from buffer \\
    Generate latent representations for buffered data at \emph{every} intermediate layer $\{\{z^M_{i, j}\}_{i=1}^{B}\}_{j=1}^{L}$\\
    Generate latent representation for current data at the \emph{last} intermediate layer
    $\{z^t_{i, L}\}_{i=1}^{B}$
    
    \tcc{HSIC Bottleneck for Rehearsal}
    Compute $\mathcal{L}_{\text{\bottleneck}}(\theta_f)$ in Eq.~(\ref{eq:HBR}) via mini-batched empirical estimation (Eq.~(\ref{eq:empirical_hsic}), same below)

    \tcc{HSIC Alignment}
    Generate projected views of the last layer representations $\{p_{\omega}(z^t_{i, L})\}_{i=1}^{L}$ and $\{p_{\omega}(z^M_{i, L})\}_{i=1}^{L}$
    
    Compute $\mathcal{L}_{\text{\interaction}}(\theta_f, \omega)$ in Eq.~(\ref{eq:interaction})
    
    \tcc{Original Rehearsal Loss}
    Compute $\mathcal{L}_{\text{CL}}(\theta)$ from the base rehearsal method

    Compute $\mathcal{L}_{\text{total}}$ in Eq.~(\ref{eq:overall}).

    Update $\theta = \{\theta_f, \theta_g\}$ and $\omega$ via back-propagation
    
}
}
 \caption{\method for Continual Learning}
 \label{alg:cp}
\end{algorithm}

\subsection{\interactionfull Loss} \label{sec:HI}
According to the complementary learning systems (CLS) theory~\cite{mcclelland1995there,kumaran2016learning}, learning task-invariant knowledge that can be shared between tasks is also critical in CL. To this end, we propose a novel \interactionfull (\interaction) loss to better capture task-invariant knowledge. 

Recall that we denote by $X^M$ the random variable for the buffered data, and $Z^M_L$ the latent representation from the last intermediate layer. We further denote by $X^t$ the random variable for the data from the current task $t$, as well as the corresponding latent representation $Z^t_L$. We define the \textit{\interactionfull} (\interaction) loss as:
\begin{equation} \label{eq:interaction}
\resizebox{.99\hsize}{!}{$
 \mathcal{L}_{\text{\interaction}}(\theta_f, \omega) = -\frac{1}{2} \left(\HSIC(Z^M_L, p_\omega(Z^t_L)) + \HSIC(p_\omega(Z^M_L), Z^t_L) \right),$}
\end{equation}
where $p_\omega$ is an multi-layer perceptron (MLP) projection head~\cite{grill2020bootstrap, chen2021exploring} for producing multiple views of the latent representation.

\textbf{Strengthening Task-Invariant Knowledge.} \interaction aims at maximizing the HSIC between the latent representations of examples from different tasks such that task-invariant knowledge can be better shared. Note that \interaction actually acts as a complementary learning objective to \bottleneck, as it encourages knowledge transfer between tasks in addition to less forgetting~\citep{hadsell2020embracing}.

\textbf{Design Choices.} The design of the \interaction loss is inspired by the Siamese representation learning paradigm~\cite{chen2021exploring}. In practice, we have empirically verified the effectiveness of the symmetrized loss and the necessity of adding the projection head. Interestingly, \interaction works quite well without the stop-gradient~\citep{chen2021exploring} operation. We suspect that this phenomenon may due to the fact that the intrinsic difference between examples from different tasks ensures latent representations do not collapse. We leave further theoretical explorations in our future work.

\begin{table*}[t]
\centering \small
\caption{Performance (in \emph{average accuracy}) comparison between \methodabbr with state-of-the-art rehearsal-based methods on  benchmark datasets with different buffer sizes and optional pre-training. All results are averaged through three independent runs.}
\begin{tabular}{l||c|c|c|c|c|c|c|c|c|c|c|c}
\toprule
\textbf{Method} & \multicolumn{3}{c|}{\textbf{Split CIFAR-10}} & \multicolumn{6}{c|}{\textbf{Split CIFAR-100}} & \multicolumn{3}{c}{\textbf{Split miniImageNet}}\\ \midrule
Pre-training & \multicolumn{3}{c|}{\xmark} & \multicolumn{3}{c|}{\xmark} & \multicolumn{3}{c|} {\textit{Tiny ImageNet}} & \multicolumn{3}{c}{\xmark}  \\ \midrule
{Upper bound} & \multicolumn{3}{c|}{\zheng{92.38}} & \multicolumn{3}{c|}{73.29} & \multicolumn{3}{c|}{75.20} & \multicolumn{3}{c}{53.55}  \\ 
{Sequential} & \multicolumn{3}{c|}{\zheng{19.67}} & \multicolumn{3}{c|}{9.29} & \multicolumn{3}{c|}{9.52} & \multicolumn{3}{c}{4.51}  \\
\midrule
\textbf{Buffer size} & $100$ & $200$ & $500$ & $200$ & $500$ & $2000$ & $200$ & $500$ & $2000$ & $1000$ & $2000$ & $5000$ \\
\midrule
ER & 36.39 & 44.79 & 57.74 & 14.35 & 19.66 & 36.76 & 18.09 & 28.25 & 43.18 & 8.37 & 16.49 & 24.17 \\ 
\rowcolor[gray]{.9} + \bf\methodabbr & 43.70 & 49.37 & 61.65 & 21.57 & 26.65 & 40.26 & 25.35 & 33.82 & 46.57 & 12.71 & 19.57 & 26.89 \\ 
X-DER-RPC & 59.29 & 65.19 & 68.10 & 35.34 & 44.62 & 54.44 & 51.40 & 57.45 & 62.46 &  25.24 & 26.38 & 29.91 \\ 
\rowcolor[gray]{.9} + \bf\methodabbr & 66.76 & 71.05 & 73.53 & 40.04 & 46.83 & 54.71 & 52.67 & 57.88 & 62.70 &  27.21 & 28.15 & 31.09  \\ 
ER-ACE & 53.90 & 63.41 & 70.53 & 26.28 & 36.48 & 48.41 & 41.85 & 48.19 & 57.34 & 17.95 & 22.60 & 27.92 \\ 
\rowcolor[gray]{.9} + \bf\methodabbr & 60.52 & 68.08 & 73.78 & 29.08 & 38.94 & 50.55 & 45.19 & 50.36 & 57.50 & 22.33 & 25.41 & 30.12  \\ 
DER++ & 57.65 & 64.88 & 72.70 & 25.11 & 37.13 & 52.08 & 26.50 & 43.65 & 58.05 &  18.02 & 23.44 & 30.43 \\ 
\rowcolor[gray]{.9} + \bf\methodabbr & {64.98} & {70.28}  & {75.94} & 31.46 & {41.86} & 53.53 & 34.10 & 50.64 & 59.02 & 24.78 & {29.37} & 34.98 \\
\bottomrule
\end{tabular}
\label{tab:table1}
\end{table*}

\subsection{Overall Objective} %
At every task, we incorporate both the \bottleneck and \interaction losses into the existing rehearsal-based learning framework. Therefore, the overall objective is:
{\footnotesize
\begin{align}
\label{eq:overall}
\begin{split}
 \mathcal{L}_{\text{total}} & = \mathcal{L}_{\text{CL}}(\theta) + \mathcal{L}_{\text{\bottleneck}}(\theta_f) + \lambda_{\text{\interaction}} \mathcal{L}_{\text{\interaction}}(\theta_f, \omega), \\
 & = \sum_{\vx, y \in \mathcal{D}_t} \ell(h_{\theta}(\vx), y) + \sum_{\vx^M, y^M \in \mathcal{M}} \ell_{\mathcal{M}}(h_{\theta}(\vx^M), y^M) + \\
 & \quad \underbrace{\lambda_{x} \sum_{j=1}^{L} \HSIC(X^M, Z^M_j) -  \lambda_{y} \sum_{j=1}^{L} \HSIC(Y^M, Z^M_j)}_{\textrm{\bottleneckfull }} - \\
 & \quad \lambda_{\text{\interaction}} \underbrace{\frac{1}{2} \left(\HSIC(Z^M_L, p_\omega(Z^t_L)) + \HSIC(p_\omega(Z^M_L), Z^t_L) \right)}_{\textrm{\interactionfull }},
\end{split}
\end{align}}

where $\lambda_{\text{\interaction}}$ is a balancing coefficient. Note that our surrogate loss terms are general enough to be combined with and further improve almost any existing rehearsal methods. The overall algorithm is described in Alg.~\ref{alg:cp}. In practice, we evaluate HSIC empirically via~(\ref{eq:empirical_hsic}) in mini-batches, following~\cite{wang2021revisiting}. Given mini-batch size $B$, the maximum intermediate dimension $d_Z = \max_j d_{Z_j}$, the computation complexity of evaluating the emiprical HSIC is $\mathcal{O}(B^2 d_Z)$~\cite{song2012feature}. Thus, the computational complexity overhead introduced by \methodabbr is $\mathcal{O}(LB^2d_Z)$.

\section{Experiments}

To evaluate the efficacy of the proposed~\methodabbr, we conduct comprehensive experiments on representative CL benchmarks, closely following the challenging class-incremental learning setting in prior works~\cite{lopez2017gradient,van2019three,wang2021learning}. We incorporate \methodabbr with multiple SOTA rehearsal-based CL methods to demonstrate performance improvement, while also comparing \methodabbr against other SOTA CL methods. We also performed an ablation study and exploratory experiments to further showcase the effectiveness of individual components.

\begin{table*}[t]
\centering \small
\caption{Performance (in \emph{average accuracy}) comparison between~\methodabbr and regularization-enhanced rehearsal methods on various benchmark datasets. ER-ACE and DER++ are representative rehearsal-based methods that all comparing methods build upon. All results are averaged through three independent runs.}
\begin{tabular}{l||c|c|c|c|c|c|c|c|c}
\toprule
\textbf{Method} & \multicolumn{3}{c|}{\textbf{Split CIFAR-10}} & \multicolumn{3}{c|}{\textbf{Split CIFAR-100}} & \multicolumn{3}{c}{\textbf{Split miniImageNet}}\\ \midrule
\textbf{Buffer size} & $100$ & $200$ & $500$ & $200$ & $500$ & $2000$ & $1000$ & $2000$ & $5000$ \\
\midrule
ER-ACE & 53.90 & 63.41 & 70.53 & 26.28 & 36.48 & 48.41 & 17.95 & 22.60 & 27.92  \\
+ sSGD & 56.26 & 64.73 & 71.45 & 28.07 & \bf 39.59 & 49.70 & 18.11 & 22.43 & 24.12  \\
+ oEwC & 52.36 & 61.09 & 68.70 & 24.93 & 35.06 & 45.59 & 19.04 & 24.32 & 29.46  \\
+ oLAP & 52.76 & 63.19 & 70.32 & 26.42 & 36.58 & 47.66 & 18.34 & 23.19 & 28.77  \\
+ OCM & 57.18 & 64.65 & 70.86 & 28.18 & 37.74 & 49.03 & 20.32 & 24.32 & 28.57 \\ 
+ LiDER & 56.08 & 65.32 & 71.75 & 27.94 & 38.43 &  50.32 & 19.69 & 24.13 & 30.00  \\
\rowcolor[gray]{.9} + \bf\methodabbr & \bf 60.52 & \bf 68.08 & \bf 73.78 & \bf 29.08 & 38.94 & \bf 50.55 & \bf 22.33 & \bf 25.41 & \bf 30.12 \\ 
\midrule
DER++ & 57.65 & 64.88 & 72.70 & 25.11 & 37.13 & 52.08 & 18.02 & 23.44 & 30.43 \\ 
+ sSGD & 55.81 & 64.44 & 72.05 & 24.76 & 38.48 & 50.74 & 16.31 & 19.29 & 24.24 \\
+ oEwC & 55.78 & 63.02 & 71.64 & 24.51 & 35.22 & 51.53 & 18.87 & 24.53 & 31.91 \\
+ oLAP & 54.86 & 62.54 & 71.38 & 23.26 & 34.48 & 50.80 & 18.91 & 25.02 & 32.78 \\
+ OCM & 59.25 & 65.81 & 73.53 & 27.46 & 38.94 & 52.25 & 20.93 & 24.75 & 31.16 \\ 
+ LiDER  & 58.43 & 66.02 & 73.39 & 27.32 & 39.25 & 53.27 & 21.58 & 28.33 & \bf 35.04 \\
\rowcolor[gray]{.9} + \bf\methodabbr & \bf{64.98} & \bf{70.28} & \bf{75.94} & \bf31.46 & \bf{41.86} & \bf 53.53 & \bf 24.78 & \bf {29.37} & 34.98 \\ 
\bottomrule
\end{tabular}

\label{tab:table2}
\end{table*}

\subsection{Experiment Setting} \label{sec:exp_setting}
\textbf{Evaluation Benchmarks.} We evaluate our \methodabbr on three representative CL benchmarks, following mainstream evaluation paradigms~\cite{zenke2017continual,buzzega2020dark,bonicelli2022effectiveness}. \\
- \textbf{Split CIFAR-10} originates from the well-known CIFAR-10~\cite{krizhevsky2009learning} dataset. It is split into 5 disjoint tasks with 2 classes per task. \\
- \textbf{Split CIFAR-100} is also a split version of CIFAR-100~\cite{krizhevsky2009learning}, which contains 10 disjoint tasks with 10 classes per task. \\
- \textbf{Split miniImageNet} is subsampled from ImageNet~\cite{deng2009imagenet} with $100$ classes. It is split into 20 disjoint tasks with 5 classes per task. \zifeng{Dataset licensing information can be found in Appendix~\ref{app:license}}.

\textbf{Comparing Methods.} We compare \methodabbr of multiple SOTA CL methods of different kinds.\\
- \textbf{Rehearsal-Based.} \methodabbr is a general framework that can be combined with almost any mainstream rehearsal-based methods. Therefore, we incorporate \methodabbr into multiple SOTA rehearsal-based methods, including \textbf{ER}~\cite{chaudhry2019tiny}, \textbf{DER++}~\cite{buzzega2020dark}, \textbf{X-DER-RPC}~\cite{boschini2022class}, and \textbf{ER-ACE}~\cite{caccia2021new}, to demonstrate its general effectiveness. \\
- \textbf{Regularization-Based.} Note that \methodabbr can also be regarded as a novel regularizer in addition to the original rehearsal-based loss. Therefore, We also compare our method with existing regularization-based techniques based on ER-ACE and DER++, including \textbf{sSGD}~\cite{mirzadeh2020understanding}, \textbf{oEWC}~\cite{schwarz2018progress}, \textbf{oLAP}~\cite{ritter2018online}, and more recent SOTA methods, \textbf{OCM}~\cite{guo2022online} and \textbf{LiDER}~\cite{bonicelli2022effectiveness}. \\
- \textbf{Reference Baselines.} For completeness, we also include the naive baseline, \textbf{Sequential}, that trains a model sequentially on tasks without any buffer, and the possible \textbf{Upper bound}, that trains the model on the union of all tasks in an i.i.d. fashion, for reference.

\textbf{Evaluation Metrics.}
We report two major metrics that are widely used in previous works~\cite{chaudhry2018efficient,lopez2017gradient,mai2021online}: \textit{Average accuracy} (higher is better) and \textit{Forgetting} (lower is better) is used to evaluate the performance of the final model trained sequentially on all tasks. The formal definition of both metrics are shown in Appendix~\ref{app:metrics}. \zifeng{Note that we include average accuracy as our main result, while keep the forgetting results and error bars in Appendix~\ref{app:forgetting} and~\ref{app:error_bars} due to space limit.}

\textbf{Experimental Details.}
We follow the standard settings in prior CL work~\cite{buzzega2020dark, bonicelli2022effectiveness} for a fair comparison. For hyperparameters of the base rehearsal methods, we refer~\cite{buzzega2020dark} for the best configurations. All methods use the same backbone model, training epochs and batch sizes for fair comparison.
The per-task training epochs are set as 50 for Split CIFAR-10/100, and 80 for Split miniImageNet. The batch sizes are set as 32, 64, and 128 for Split CIFAR-10, Split CIFAR-100, and Split miniImageNet, respectively. We adopt ResNet-18~\cite{he2016deep} without any pre-training for Split CIFAR-10 and Split CIFAR-100. Additionally, we experiment on Split CIFAR-100 using ResNet-18 pre-trained on Tiny ImageNet as reported in \cite{bonicelli2022effectiveness}. For Split miniImageNet, EfficientNet-B2~\cite{tan2019efficientnet} without pre-training is used. For the projection head, we use a $128$-$512$-$128$ MLP and a $352$-$1408$-$352$ MLP for ResNet-18 and EfficientNet-B2, respectively. To implement the empirical HSIC, we adopt the commonly used Gaussian kernel with $\sigma = 5$, following the recommendations by~\citet{wang2021revisiting}. \zifeng{Additional details about the selection of balancing coefficients are reported in Appendix~\ref{app:coefficient}.}

For the comparing methods, we either directly take existing results reported in~\citet{bonicelli2022effectiveness}, or reproduce the experiment results using the suggested hyperparameters from their original papers. 

\textbf{Computing Resources.} All experiments are conducted on a single Tesla V100 GPU with 32GB memory.%

\subsection{Comparison with Rehearsal Methods}
Table \ref{tab:table1} presents our evaluation results comparing~\methodabbr with multiple SOTA rehearsal-based models. We can see that DualHSIC can consistently improve the performance of all base methods in almost all evaluated scenarios, in terms of both average accuracy and forgetting (Appendix~\ref{app:forgetting}). In particular, the maximum performance gain of DualHISC is $7.6\%$ across all datasets and buffer sizes. Interestingly, we observe the largest performance gap when buffer size is small in Table~\ref{tab:table1}, while similar trend is revealed in Figure~\ref{fig:intro} as well. This observation actually confirms the effectiveness of~\methodabbr under more challenging scenarios. As suggested by~\citet{prabhu2020gdumb}, when buffer size is large, the data imbalance issue between buffered and current data is naturally mitigated. Moreover, we observe that the effectiveness of \methodabbr is orthogonal to pre-trainig, showing the potential that \methodabbr 
can be useful in real-world scenarios that often involves learning from a pre-trained model~\citep{wang2021learning}.

\subsection{Comparison with Regularization-Enhanced Rehearsal Methods}
To further demonstrate the effectiveness of~\methodabbr, we compare~\methodabbr against regularization-enhanced rehearsal methods by combining existing regularization techniques and replay strategies including ER-ACE and DER++. We show the experiment results in Table \ref{tab:table2}. \methodabbr outperforms almost all regularization techniques on all benchmarks with various buffer sizes, by at most $6.5\%$ margin. Even when \methodabbr is outperformed, its gap with the top performer is minimal ($<0.6\%$). Similarly, we observe the clear advantage of~\methodabbr at the small buffer regime. Note that sSGD, oEwC and oLAP do not specifically consider buffered data, while OCM and LiDER do not explicitly consider the inter-task relationship as~\methodabbr does, which may account for the larger performance gap when buffer size is small.

\begin{figure}[t]
    \centering
    \includegraphics[width=0.89\columnwidth]{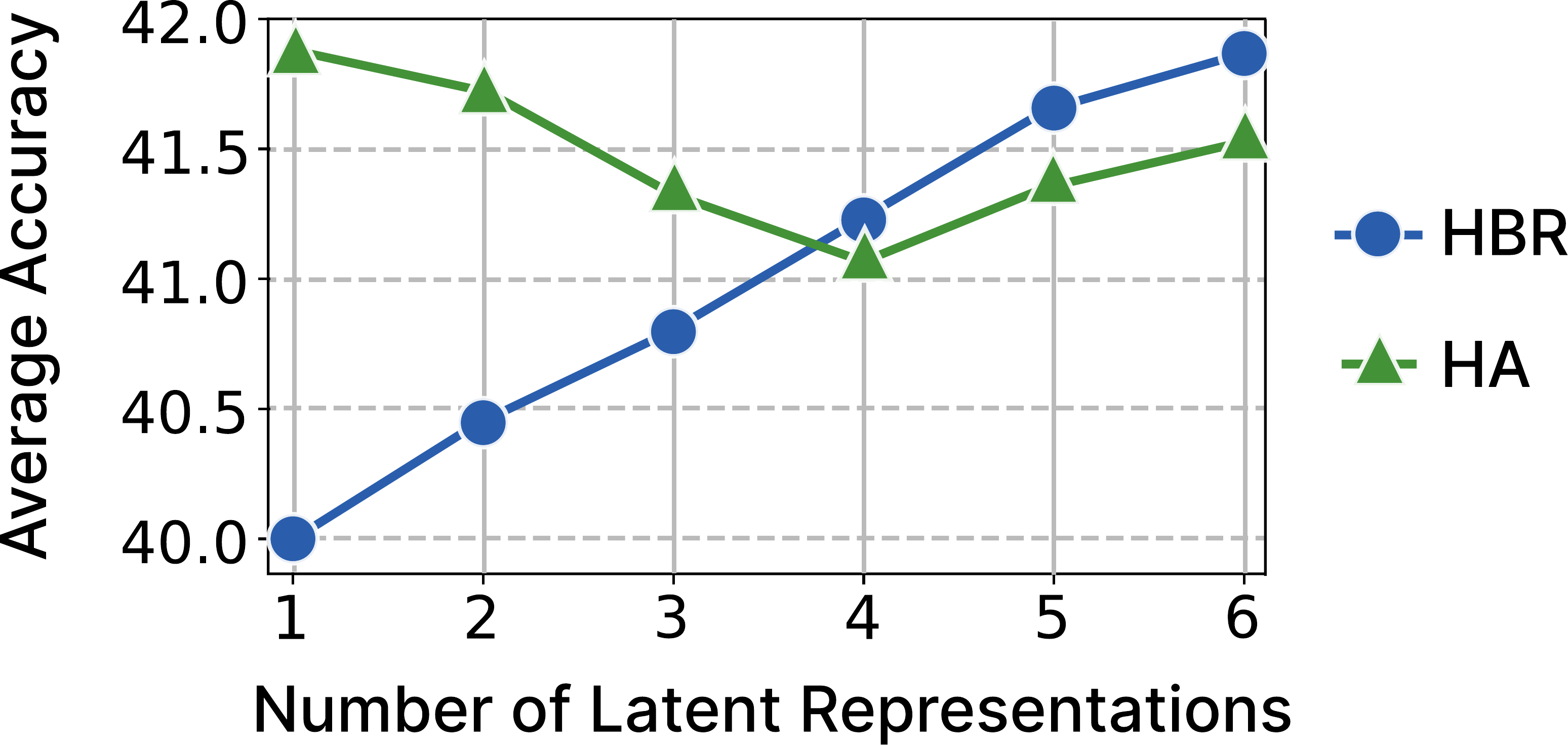}
    \caption{Effectiveness of \bottleneck and \interaction w.r.t. number of latent representations included. \bottleneck gets increasingly better performance with more latent representations, while \interaction gets the best performance with a single latent representation.}
    \label{fig:layer_vs_acc}
\end{figure}

\begin{figure*}[t]
    \centering
    \includegraphics[width=0.99\textwidth]{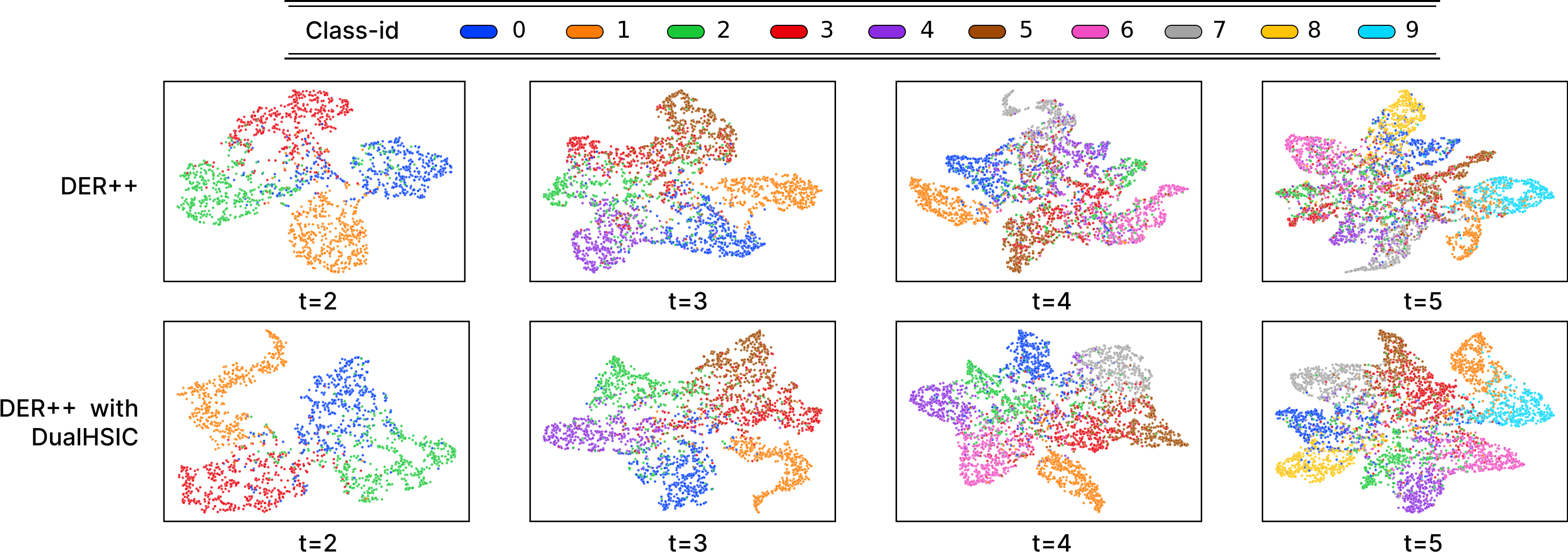}
    \caption{Comparison of tSNE visualization of the last latent representation between DER++ vs. DER++ with \methodabbr. We show the progress of tSNE embeddings starting from task $2$, since task $1$ can be regarded as trivial supervised learning with no catastrophic forgetting issue. Classes are split into 5 tasks based on their label ID, \ie, 2 consecutive classes are assigned to the same task. \methodabbr improves DER++ with more separable embeddings and less inter-class confusion at later tasks.}
    \vspace{-0.1cm}
    \label{fig:visualization}
\end{figure*}

\begin{table}[t]
\centering 
\caption{Ablation study on Split CIFAR-100 with 500 buffer size, $\mathcal{L}_{\text{CL}}$ is based on DER++. $\operatorname{H}(\cdot; \cdot)$ represents $\operatorname{HSIC}(\cdot; \cdot)$. ``Proj.'' and ``Sym.'' are abbreviations for using projection head $p_{\omega}$ and using symmetric loss, respectively.}
    \centering 
    \scalebox{0.8}{
    \begin{tabular}{ccccc|c}
    \toprule 
     \multirow{2}{*}{$\mathcal{L}_{\text{CL}}$}  & \multicolumn{2}{c}{$\mathcal{L}_{\text{\bottleneck}}$} & \multicolumn{2}{c|}{$\mathcal{L}_{\text{\interaction}}$} & \multirow{2}{*}{Class-IL Acc ($\uparrow$)} \\
     & $\operatorname{H}(X; Z)$ & $\operatorname{H}(Y; Z)$ & Proj. & Sym. & \\
    \midrule
    \cmark& \xmark & \xmark & \xmark & \xmark & 37.13  \\
    \cmark& \cmark & \xmark & \xmark & \xmark & 39.60  \\
    \cmark & \cmark & \cmark & \xmark & \xmark & 40.79 \\
    \cmark & \cmark & \cmark &  \cmark & \xmark & 41.68 \\
    \cmark & \cmark & \cmark &  \cmark & \cmark &\bf 41.86 \\
    \bottomrule
    \end{tabular}
    }
    \label{table:ablation}
    \vspace{-0.4cm}
\end{table}

\subsection{Effectiveness of Core Designs} \label{sec:ablation}
    
\textbf{Ablation Study.} We perform a comprehensive ablation study by evaluating the contribution of each component in \methodabbr on Split CIFAR-100, with buffer size equal to 500, and the results are shown in Table \ref{table:ablation}. \textit{In summary, all components of~\methodabbr contributes to the final performance improvement}.
Firstly, introducing the term $\operatorname{H}(X; Z)$ alone can improve the the accuracy by 2.5\%. The rationale behind this is that minimizing the HSIC between $X$ and $Z$ can help reduce the noisy information contained within the latent representation w.r.t. the input, thus mitigating the catastrophic forgetting problem. From the third row we can see that we further improve the accuracy by 1.2\% by incorporating $\operatorname{H}(Y; Z)$ into the loss, which helps preserve the discriminative information useful for classification. We can see the collaborative performance of \bottleneck as a whole, similar to the observation made by~\citet{wang2021revisiting}. The forth and fifth row in Table~\ref{table:ablation} show the necessity of the projection head $p_\omega$ and the symmetric loss term. We have empirically found that each of them can further improve the final performance which can support \methodabbr to learn task-invariant knowledge better.

\textbf{Multi-Layered vs. Single-Layered Loss.} In our final formulation of~\methodabbr, we use multi-layered loss for \bottleneck and single-layered loss for \interaction. To validate this specific design choice, we present how an increasing number of latent representations obtained from multiple layers affects the final performance, on both \bottleneck and \interaction in Figure~\ref{fig:layer_vs_acc}. Experiment details are included in Appendix~\ref{app:multi_single_layer}. Interestingly, \bottleneck performs better with more latent representations from multiple layers, while \interaction does not gain benefits from adding more representations. One possible reason may be that mitigating task-interference is essentially a harder task than encouraging sharing of task-invariant features, considering the data imbalance issue between buffered and current data. Therefore adding multiple levels of intermediate supervision strengthens the performance of \bottleneck. On the other hand, note that the multi-layered version of \interaction requires independent projection heads at each intermediate layer, due to the dimension difference. When the performances are close, we choose the single-layered with minimal overhead for \interaction.

\textbf{Visualization of Latent Representation.} We compare t-SNE~\cite{van2008visualizing} visualization of the last latent representation between DER++ with and without \methodabbr during CL on Split CIFAR-10 with \zifeng{200} buffer size. The t-SNE visualization at the end of task $2$ to $5$ is shown in Figure~\ref{fig:visualization}, where different colors represent different class labels. We observe that the latent representations from earlier tasks are better separated in the embedding space when trained with \methodabbr additionally. For example, in the upper row (DER++), class 2 and 3 from the second task are getting scattered and largely overlapped with classes from other tasks at the $4$-th and $5$-th tasks. While in the lower row (DER++ with~\methodabbr), the class separation are better maintained throughout the CL learning process, thanks to the synergy between \bottleneck and \interaction.

\vspace{-0.25cm}
\section{Conclusion}
In this paper, we propose \methodabbr, a general method for continual learning that can mitigate inter-task interference and extract task-invariant knowledge at the same time. It has two key components based on the so-called Hilbert-Schmidt independence criterion (HSIC): \bottleneckfull (\bottleneck) and \interactionfull (\interaction). We conduct comprehensive experiments and an ablation study to show that \methodabbr can be seamlessly plugged into a wide range of SOTA rehearsal-based methods and consistently improve the performance under different settings. Moreover, We recommend \methodabbr as a starting point for future research on the effectiveness of HSIC in CL.

\section*{Potential Negative Societal Impacts}
\methodabbr is a novel and effective CL method to enhance various rehearsal-based methods and has great practical potential. However, we should be cautious of the potential negative societal impacts it might lead to. For example, as \methodabbr can be integrated into almost any rehearsal-based frameworks, we should always double check and mitigate the fairness and bias~\citep{mehrabi2021survey} issues that existed in the base model before we further deploy \methodabbr, in case such issues propagate. Moreover, when applying \methodabbr to privacy-sensitive~\citep{al2019privacy} applications, we need to ensure the buffered data are well-anonymized to prevent privacy breach. %
In summary, we would recommend to analyze and prepare possible solutions to potential negative societal impacts in detail, before deploying \methodabbr in real-world applications.

\section*{Limitations}
Although \methodabbr is a pioneering work that first introduces HSIC into CL for reducing inter-task interference and learning better task-invariant knowledge, we would still like to discuss the current limitations of \methodabbr. First, \methodabbr aims at improving widely-adopted rehearsal-based methods. When rehearsal buffer is not allowed, the formulation of \methodabbr potentially needs to be revised to work. Second, we motivate \methodabbr intuitively and demonstrate the effectiveness of \methodabbr empirically by comprehensive experiments. However, theoretical foundation is still under exploration to strictly link HSIC with catastrophic forgetting. We would like to treat current limitations of our work as interesting research directions and topics for our future work.

\newpage
\appendix
\onecolumn

\section*{Checklist}

\begin{enumerate}
\item Have you read the publication ethics guidelines and ensured that your paper conforms to them? Yes.
\item Did you discuss any potential negative societal impacts (for example, disinformation, privacy, fairness) of your work? Yes, please see the Potential Negative Societal Impacts section.
\item If you are using existing assets (e.g., code, data, models) or curating/releasing new assets...
\begin{itemize}
    \item If your work uses existing assets, did you cite the creators and the version? Yes.
    \item Did you mention the license of the assets? Yes, please see Apendix~\ref{app:license}.
    \item Did you include any new assets either in the supplemental material or as a URL? No.
\end{itemize}
\item Do the main claims made in the abstract and introduction accurately reflect the paper's contributions and scope? Yes.
\item Did you describe the limitations of your work? Yes, please see the Limitations section.
\item If you ran experiments...
\begin{itemize}
    \item Did you include the code, data, and instructions needed to reproduce the main experimental results (either in the supplemental material or as a URL, in an anonymized way at the review time)? Yes, please see the supplementary materials.
    \item Did you specify all the training details (e.g., data splits, hyperparameters, how they were chosen)? Yes, please see Section~\ref{sec:exp_setting} and Appendix~\ref{app:coefficient}.
    \item Did you report error bars (e.g., with respect to the random seed after running experiments multiple times)? Yes, please see Appendix~\ref{app:error_bars}.
    \item Did you include the amount of compute and the type of resources used (e.g., type of GPUs, internal cluster, or cloud provider)? Yes, please see Section~\ref{sec:exp_setting}.
\end{itemize}

\end{enumerate}

\newpage
\section{Dataset Licensing Information} \label{app:license}
\begin{itemize}
    \item CIFAR-10 and CIFAR-100 are licensed under the MIT license.
    \item miniImageNet is licensed under the CC0: Public Domain license.
\end{itemize}

\section{Evaluation Metrics} \label{app:metrics}
We define the average accuracy and forgetting following~\cite{lopez2017gradient}.
Let $S_{t, \tau}$ be the classification accuracy on the $\tau$-th task after training on the $t$-th task. When the model has been trained sequentially on the first $t$ tasks, the \emph{average accuracy} ($A_t$) and \emph{forgetting} ($F_t$) can be computed as follows:
\begin{equation*}
\begin{aligned}
&A_{t}=\frac{1}{t} \sum_{\tau=1}^{t} S_{t, \tau}\\
&F_{t}=\frac{1}{t-1} \sum_{\tau=1}^{t-1} \max _{\tau^{\prime} \in\{1, \cdots, t-1\}}\left(S_{\tau^{\prime}, \tau}-S_{t, \tau}\right)
\end{aligned}
\end{equation*}

\section{Additional Experiment Details and Results}
\subsection{Balancing Coefficients} \label{app:coefficient}
As suggested by~\citep{wang2021revisiting}, the balancing coefficients of HSIC-related terms should roughly follow the rule-of-thumb that the balanced losses should be the same scale as the original loss. We follow the recommendation as starting points for searching the optimal $\lambda_x$, $\lambda_y$ and $\lambda_{\text{\interaction}}$ on the validation set, which contains $20\%$ random sampled from the training set. The final sets of balancing coefficients for different datasets are as follows:
\begin{table}[h]
\centering
\caption{Optimal balancing coefficients.}
\begin{tabular}{c|ccc}
\toprule
Dataset & $\lambda_x$ & $\lambda_y$ & $\lambda_{\text{\interaction}}$ \\ \midrule
Split CIFAR-10 & 0.001 & 0.05 & -0.75 \\ 
\hline
 Split CIFAR-100 & 0.001 & 0.05 & -0.75 \\ 
\hline
Split miniImageNet & 0.001 & 0.1 & -0.75 \\ 
\bottomrule
\end{tabular}
\end{table}

\subsection{Error Bars} \label{app:error_bars}
We report the corresponding error bars of Table~\ref{tab:table1} in the main text in Table~\ref{tab:app-error-bar}. 

\begin{table*}[t]
\centering \small
\caption{Error bars of Table~\ref{tab:table1}. `-' means the result is taken from existing work.}
\begin{tabular}{l||c|c|c|c|c|c|c|c|c|c|c|c}
\toprule
\textbf{Method} & \multicolumn{3}{c|}{\textbf{Split CIFAR-10}} & \multicolumn{6}{c|}{\textbf{Split CIFAR-100}} & \multicolumn{3}{c}{\textbf{Split miniImageNet}}\\ \midrule
Pre-training & \multicolumn{3}{c|}{\xmark} & \multicolumn{3}{c|}{\xmark} & \multicolumn{3}{c|} {\textit{Tiny ImageNet}} & \multicolumn{3}{c}{\xmark}  \\ \midrule
{Upper bound} & \multicolumn{3}{c|}{\zheng{92.38}} & \multicolumn{3}{c|}{73.29} & \multicolumn{3}{c|}{75.20} & \multicolumn{3}{c}{53.55}  \\ 
{Sequential} & \multicolumn{3}{c|}{\zheng{19.67}} & \multicolumn{3}{c|}{9.29} & \multicolumn{3}{c|}{9.52} & \multicolumn{3}{c}{4.51}  \\
\midrule
\textbf{Buffer size} & $100$ & $200$ & $500$ & $200$ & $500$ & $2000$ & $200$ & $500$ & $2000$ & $1000$ & $2000$ & $5000$ \\
\midrule
ER & 1.23 & 1.64 & 0.97 & 0.64 & 1.56 & 0.94 & 1.33 & 0.69 & 2.00 & 1.11 & 0.93 & 1.46 \\ 
\rowcolor[gray]{.9} + \bf\methodabbr & 0.69 & 1.51 & 0.33 & 1.02 & 0.63 & 0.78 & 0.55 & 0.87 & 1.44 & 0.05 & 0.58 & 0.83 \\ 
X-DER-RPC & 1.86 & 0.32 & 1.39 & 2.26 & - & - & 2.17 & - & - &  1.96 & - & - \\ 
\rowcolor[gray]{.9} + \bf\methodabbr & 1.07 & 0.67 & 0.89 & 0.46 & 1.06 & 0.41 & 1.81 & 1.04 & 0.57 &  0.42 & 1.30 & 1.58  \\ 
ER-ACE & 0.88 & 0.79 & 0.99 & 1.12 & - & - & 0.83 & - & - & 0.43 & - & - \\ 
\rowcolor[gray]{.9} + \bf\methodabbr & 1.04 & 1.22 & 1.17 & 1.08 & 0.35 & 1.57 & 0.21 & 0.43 & 1.00 & 1.49 & 0.64 & 0.72  \\ 
DER++ & 1.90 & 0.91 & 0.53 & 1.32 & - & - & 0.96 & -& - & 1.42 & -& - \\ 
\rowcolor[gray]{.9} + \bf\methodabbr & {0.46} & {1.64}  & {0.06} & 0.53 & 0.66 & 1.42 & 0.85 & 0.81 & 1.40 & 1.33 & {0.87} & 1.29 \\
\bottomrule
\end{tabular}
\label{tab:app-error-bar}
\end{table*}

\subsection{Forgetting} \label{app:forgetting}
We report the corresponding forgetting metric of Table~\ref{tab:table1} in the main text in Table~\ref{tab:app-forgetting}.

\begin{table*}[t]
\centering \small
\caption{Performance (in \emph{forgetting}, lower is better) comparison between \methodabbr with state-of-the-art rehearsal-based methods on  benchmark datasets with different buffer sizes and optional pre-training. All results are averaged through three independent runs.}
\begin{tabular}{l||c|c|c|c|c|c|c|c|c|c|c|c}
\toprule
\textbf{Method} & \multicolumn{3}{c|}{\textbf{Split CIFAR-10}} & \multicolumn{6}{c|}{\textbf{Split CIFAR-100}} & \multicolumn{3}{c}{\textbf{Split miniImageNet}}\\ \midrule
Pre-training & \multicolumn{3}{c|}{\xmark} & \multicolumn{3}{c|}{\xmark} & \multicolumn{3}{c|} {\textit{Tiny ImageNet}} & \multicolumn{3}{c}{\xmark}  \\ \midrule
{Upper bound} & \multicolumn{3}{c|}{\zheng{92.38}} & \multicolumn{3}{c|}{73.29} & \multicolumn{3}{c|}{75.20} & \multicolumn{3}{c}{53.55}  \\ 
{Sequential} & \multicolumn{3}{c|}{\zheng{19.67}} & \multicolumn{3}{c|}{9.29} & \multicolumn{3}{c|}{9.52} & \multicolumn{3}{c}{4.51}  \\
\midrule
\textbf{Buffer size} & $100$ & $200$ & $500$ & $200$ & $500$ & $2000$ & $200$ & $500$ & $2000$ & $1000$ & $2000$ & $5000$ \\
\midrule
ER & 55.90 & 44.46 & 38.15 & 70.17 & 63.92 & 46.56 & 70.33 & 58.50 & 30.02 & 63.55 & 54.14 & 41.32 \\ 
\rowcolor[gray]{.9} + \bf\methodabbr & 47.18 & 39.53 & 33.91 & 64.61 & 56.23 & 39.05 & 63.53 & 54.20 & 28.83 & 51.03 & 44.58 & 37.97 \\ 
X-DER-RPC & 30.16 & 23.16 & 17.48 & 41.58 & 31.84 & 17.01 & 22.68 & 16.86 & 12.07 & 49.93 & 38.33 & 28.29  \\ 
\rowcolor[gray]{.9} + \bf\methodabbr & 26.38 & 21.56 & 17.08 & 35.82 & 27.59 & 12.02 & 15.33 & 11.93 & 11.10 & 33.51 & 25.94 & 21.60  \\ 
ER-ACE & 22.76 & 18.30 & 14.96 & 50.63 & 38.21 & 27.90 & 39.42 & 31.84 & 25.48  & 29.82 & 23.74 & 19.72 \\ 
\rowcolor[gray]{.9} + \bf\methodabbr & 18.53 & 16.30 & 12.05 & 43.91 & 34.52 & 28.02 & 32.70 & 27.36 & 26.08 & 27.87 & 24.37 & 19.56  \\ 
DER++ & 40.25 & {30.06} & {21.85} & 62.92 & 49.80 & 31.10 & 71.26 & 48.72 & 29.65 & 63.40 & 46.69 & 37.11  \\ 
\rowcolor[gray]{.9} + \bf\methodabbr & {32.52} & {24.57}  & {17.73} & 54.96 & {45.81} & 27.52 & 65.94 & {46.73} & 26.40 & 48.16 & {34.44} & 25.56 \\
\bottomrule
\end{tabular}
\label{tab:app-forgetting}
\end{table*}

\subsection{Details of Multi-Layer vs. Single-Layered Loss} \label{app:multi_single_layer}

We design the experiment based on our final model, \ie, \bottleneck is calculated using every latent representation, and \interaction only uses the final latent representation. To verify the design choice, we conduct experiments using DER++ with \methodabbr and vary the number of latent representations added to both terms. We use ResNet18 as the backbone on Split CIFAR-100 with 500 buffer size. Following~\citep{wang2021revisiting}, we treat a ResNet basic block as a whole for generating latent representations. Therefore, we can get $6$latent representations in total. Specifically, for \bottleneck, we increase the number of latent representations in a forward fashion, \ie, starting from only adding the very first representation to adding all representations. For \interaction, we increase the number of latent representations in a backward fashion, \ie, starting from only adding the very last representation to adding all representations.

\subsection{Add \bottleneck to Current Data} \label{app:cur_data}
We conduct an exploratory study to validate our asynchronous consolidation strategy of \bottleneck discussed in Section~\ref{sec:HBR}. Specifically, we compare (1) adding \bottleneck only to the buffer, (2) adding \bottleneck to only current data, (3) adding \bottleneck to both buffer and current data, using DER++ with \methodabbr on Split CIFAR-100 with 500 bufer size. We clearly observe from Table~\ref{tab:table5} that only adding \bottleneck to the buffered data yields the best performance. On the contrary, adding \bottleneck to current data hinders the learning of the current task, possibly due to the fact that \bottleneck not only removes noisy information, but also mistakenly removes useful task-specific information for the current task.
\begin{table}[h]
\centering
\caption{Exploration of adding \bottleneck to current or buffered data.} \label{tab:table5}
\begin{tabular}{cc|cc}
\toprule
Buffer & Current & Average Acc. & Forgetting \\ \midrule
\xmark& \xmark & 37.13 & 48.72 \\
\cmark& \xmark & 41.86 & 36.96 \\ 
\xmark& \cmark & 36.64 & 45.55 \\ 
\cmark& \cmark & 38.78 & 43.68 \\ 
\bottomrule
\end{tabular}
\end{table}

\end{document}